\documentclass[runningheads]{llncs}
\usepackage{graphicx}
\usepackage{amsmath}
\usepackage{array}
\usepackage{algorithm2e}
\RestyleAlgo{ruled}

\usepackage{todonotes}
\usepackage[skip=3mm]{caption}
\usepackage{caption}
\usepackage{url}
\usepackage{hyperref}
\usepackage{subcaption}

\newsavebox{\twosubbox}
\begin{document}
\title{Potential-based Reward Shaping in Sokoban}
%
%
\author{
Zhao Yang\inst{1}
\and
Mike Preuss\inst{2}
\and
Aske Plaat\inst{3}
}
\authorrunning{Z. Yang et al.}
%
\institute{LIACS, Leiden University, the Netherlands \\
\email{z.yang@liacs.leidenuniv.nl}
\and
LIACS, Leiden University, the Netherlands \\
\email{m.preuss@liacs.leidenuniv.nl}
\and
LIACS, Leiden University, the Netherlands \\
\email{aske.plaat@gmail.com}}
\maketitle              
\begin{abstract}
Learning to solve sparse-reward reinforcement learning problems is difficult, due to the lack of guidance towards the goal.  But in some problems,  prior knowledge can be used to augment the learning process. Reward shaping is a way to incorporate prior knowledge into the original reward function in order to speed up the learning. While previous work has investigated the use of expert knowledge to generate potential functions, in this work, we study whether we can use a search algorithm(A*) to automatically generate a potential function for reward shaping in Sokoban, a well-known planning task. The results showed that learning with shaped reward function is  faster than learning from scratch. Our results  indicate that distance functions could be a suitable function for Sokoban. This work demonstrates the possibility of solving multiple instances with the help of reward shaping. The result can be compressed into a single policy, which can be seen as the first phrase towards training a general policy that is able to solve unseen instances.\footnote{We open-sourced all the code we used. It can be found, after the review, at\\ 
\url{https://anonymous.org/blind-review}
}

\end{abstract}

\section{Introduction}
\begin{figure}[!ht]
    \centering
    \includegraphics[scale=0.5]{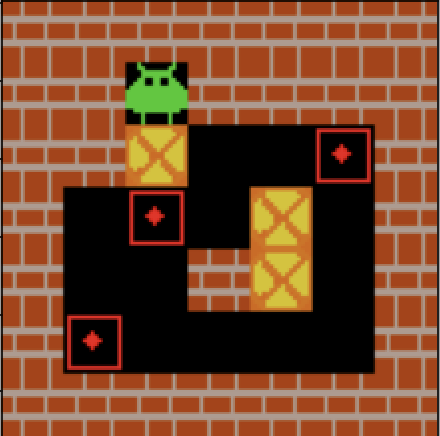}
    \caption{An example of 3-boxes Sokoban instance.}
    \label{fig:example}
\end{figure}

Sokoban is a well-known puzzle game that is often used as a benchmark for evaluating reinforcement learning (RL) agents~\cite{guez2019investigation,hamrick2021on}.  Sokoban is a sparse reward task, furthermore, it suffers from dead-ends: one bad action can render the whole instance unsolvable. Sokoban is a deceptively simple puzzle. RL agents struggle to learn a behavior policy, unless they used planning as part of their learning effort.  A simple example from~\cite{SchraderSokoban2018} is shown in Fig.~\ref{fig:example}. Although this example has only three boxes, it might already be  challenging to solve for humans.

Human problem solving used heuristics, rules of thumb that are based on experience, that work most of the time, but not always. Heuristics usually increase our ability to solve problems greatly.   Reward shaping~\cite{dorigo1994robot,Ng99policyinvariance} is proposed for incorporating prior (heuristic) knowledge to accelerate learning in RL. It reshapes the original reward function by adding another reward function which is formed by prior knowledge in order to get an easy-learned reward function, that is often also more dense. Examples of prior knowledge are heuristics from context structures, demonstrations from experts, etc. 

In this paper, we show that reward shaping can be applied to sparse reward tasks for faster learning. More accurately, we choose the distance function  (automatically provided by the A* search algorithm) as the potential function, and subsequently performed potential-based reward shaping in Sokoban. Our results demonstrate that learning with shaped reward functions outperforms learning from scratch by a large margin. In contrast to neural networks, that are able to generalize to unseen tasks but require much training data, our reward shaping can be seen as the first step towards the final goal that aims to train an agent which is able to solve multiple unseen new Sokoban instances. With reward shaping, the ability of solving multiple instances is  compressed into a single behavior policy without extensive training.

The paper is structured as follows: first we briefly describe related work  in section~\ref{related}; then details about our method are provided in section~\ref{methods}; followed by experimental design and results; lastly, we discuss limitations and potential future works of this paper, then draw  conclusions in section~\ref{conclusion}.

\section{Related Work}
\label{related}
Reinforcement learning (RL) algorithms are used to solve decision making problems which could be formed into Markov Decision Process (MDPs), and they train policies by interacting with environments~\cite{plaat2020learning,sutton2018introduction}. Recently, RL achieves super human performance in the board game Go~\cite{silver2016mastering}, Atari games~\cite{badia2020agent57} and StarCraft~\cite{vinyals2019grandmaster}. In this section, we will briefly describe related work about both potential-based reward shaping and Sokoban. 
\subsection{Reward Shaping}
Reward shaping offers a way to add useful information to the reward function of the original MDP. By reshaping, the original sparse reward function will be denser and is more easily-learned. The heuristics can come from different sources, such as demonstrations either from human or another RL agent~\cite{brys2015reinforcement,hester2018deep}, or expert's guidance, etc.  

The optimal policy is determined by the reward function, small transformations of the reward function might cause intractable problems~\cite{randlov1998learning}. Ng. et al. proved that by following the potential-based reward shaping, the optimal policy will be invariant~\cite{Ng99policyinvariance}. The original reward function $R$ is augmented by another reward function $F$, shown in Eq.~\ref{eq1} and if and only if $F$ is the subtraction between a function $\phi$ of the next state $s'$ and the current state $s$ then the optimal policy will keep unchanged. The function $\phi$ is called potential function. Examples of good potential function could be Manhattan distance in navigation tasks or pre-trained state value functions, etc.
\begin{equation} \label{eq1}
\begin{split}
R'(s,a,s') & = R(s,a,s')+F(s,a,s') \\
 & = R(s,a,s')+\phi(s')-\phi(s) \\
\end{split}
\end{equation}

Brys et al. extracted potential function from demonstrations by checking if agent's state-action pairs are in demonstrations or not and apply it to Cart Pole and Mario~\cite{brys2015reinforcement}. Hussein et al. trained a neural network from demonstrations as the potential function and added it to the original reward function of DQN in grid navigation tasks\cite{hussein2017deep}. Grzes provided more insights and analysis for potential-based reward shaping and extended it to multi-agent RL scenario~\cite{grzes2017reward}. While most previous methods have focused on extracting potential functions from expert demonstrations, we investigate whether potential functions can also be extracted from a search. In our case, we use the distance function which is provided by the A* search algorithm as the potential function.

\subsection{Sokoban}
Sokoban is a  challenging puzzle game and has been proved to be PSPACE-complete~\cite{culberson1997sokoban} and NP-hard~\cite{dor1999sokoban} problem. It also plays an  important role in benchmarking RL agents. Many  models are proposed to solve Sokoban. Both model-based methods~\cite{guez2018learning,hamrick2021on,weber2017imagination}, as well as  model-free methods can  reach competitive performance~\cite{guez2019investigation}. Curriculum learning has been used to solve  a difficult Sokoban instance~\cite{feng2020novel}. The works mentioned above try to solve Sokoban using special-designed models, while we are focusing on using general  reward shaping techniques to speed up the learning. 

Fine-tuning pre-trained models is helpful in accelerating learning in Sokoban~\cite{yang2021transfer}. Reward shaping was applied to a single simple Sokoban instance by interacting with human experts in~\cite{raza2016reward} to speed up the learning. In our work, we demonstrate potential-based reward shaping over many Sokoban instances range from 1-box to 3-boxes, where no human expert involved.

\section{Methods}
\label{methods}
In this section, we will explain the methods and techniques that we used. We report the problem model, the heuristic that was used in A*, and details about the reward shaping.

\subsection{Reinforcement Learning}
MDP is short for Markov Decision Process, and it models decision making problems into a 4-tuple, $\langle S, A, R(s,a,s'), P(s,a,s')\rangle$. In our paper, we follow the MDP notation proposed in~\cite{sutton2018introduction}. $S$ is a set of states $s$ called the state space, it will be different states in Sokoban instance in our case. $A$ is a set of actions $a$ called the action space, in our case it will contain all actions that the agent can take (no operation, going up/down/left/right). $R(s,a,s')$ is a reward function that determines immediate rewards that the agent will get after performs an action $a$ which leads the agent from the current state $s$ to the next state $s'$. $P(s,a,s')$ is the probability that action $a$ leads the agent from the current state $s$ to the next state $s'$. Reinforcement learning methods solve MDP using data($s,a,s',R(s,a,s')$) collected by interacting with the environment to train a policy aims to maximize the accumulated reward shown in Eq.~\ref{eq:r},
\begin{equation}
R=\sum\limits_{t=0}^{\infty} \gamma^t r_{t}    
\label{eq:r}
\end{equation}where $\gamma \in [0,1]$ is the discount factor and $r_t$ is the immediate reward the agent gets in time step $t$.

We use the RL algorithm A2C to train the agent to learn to solve Sokoban. The policy is represented by a neural network and the architecture of the neural network we are using for experiments is the same as the architecture used in~\cite{yang2021transfer}, which consists of 3 convolutional layers and 2 fully-connected layers. All hyper-parameters of A2C are also kept the same as described in~\cite{yang2021transfer}. More details can be found in the Appendix~\ref{nn}.

\subsection{A* Heuristics}
A* is a heuristic search algorithm, it extends the Dijkstra's algorithm by adding heuristics. The heuristic we used in our case is the overall Manhattan distance between untargeted boxes and goals\footnote{The implementation we are using is from \url{https://github.com/KnightofLuna/sokoban-solver}}, formula shown in Eq~\ref{h}.
\begin{equation}
  h(s) = \sum_{b\in B,t\in T} (|x_b-x_t|+|y_b-y_t|)
  \label{h}
\end{equation}, where $(x_b, y_b)$ is the location of boxes while $(x_t, y_t)$ is the location of targets, $h$ is the heuristic for the current state $s$. $B$ is all boxes which are not on targets yet and $T$ is all targets where there are no boxes on. If a Sokoban instance is solvable, A* will return the solution otherwise it will return nothing. As such, it could also be used to check the solvability of a Sokoban instance. 

\subsection{Reward Shaping}
Values of potential functions of states should be higher if states are 'better' and vice versa. For this reason  the \textbf{minus} of the distance function is used as the potential function in our case. The distance function will take the current state as input, and output how many steps the agent needs to take towards the goal state.

The shaped reward function will be(shown in Eq.~\ref{eq2}):

\begin{equation} \label{eq2}
\begin{split}
R'(s,a,s') & = R(s,a,s')+F(s,a,s') \\
 & = R(s,a,s')+\phi(s')-\phi(s) \\
 & = R(s,a,s')-d(s')+d(s)
\end{split}
\end{equation} where $s$ is the current state while $s'$ is the next state, and $a$ is the  action that leads $s$ to $s'$. $\phi$ is the potential function and $d(s)$ is the distance function from the current state $s$ to the goal state. In our case, we use the  A* search algorithm\footnote{\url{https://en.wikipedia.org/wiki/A*_search_algorithm}} to provide the distance information. 

\begin{figure}[!ht]
    \centering
    \includegraphics[scale=0.4]{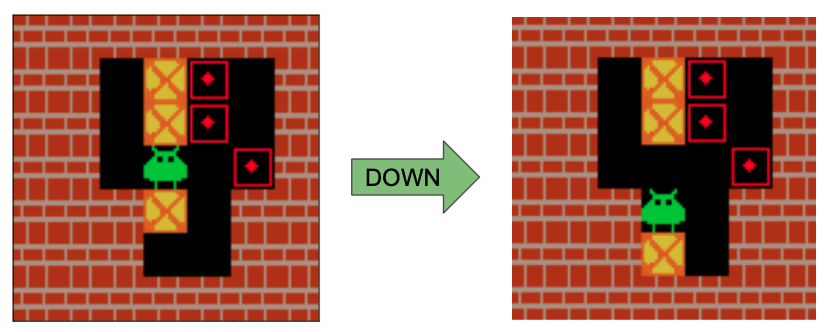}
    \caption{How an unsolvable situation happens.}
    \label{fig:irrversible}
\end{figure}

In Sokoban,  some actions can lead to unsolvable situations. An example is shown in Fig.~\ref{fig:irrversible}. A box is pushed into the corner and it is not possible to pull it back. The instance has become completely unsolvable. Then a natural question is what distance we should assign to states which are  unsolvable. The algorithm, however, can still get some rewards by learning sub-optimal policies, such as pushing one of the boxes onto one of the targets. In order  not to break the sub-optimal policy invariance, we don't shape the reward function and just keep the original reward function after the instance has become unsolvable. To conclude, our shaped reward function is shown in Eq.~\ref{overall}.
\begin{equation}
  F(s,a,s') =
    \begin{cases}
      \phi(s)-1=-d(s)-1 & \text{if $s$ is solvable and $s'$ is unsolvable}\\
      0 & \text{if both $s'$ and $s$ are unsolvable}\\
      \phi(s')-\phi(s)=-d(s')+d(s) & \text{otherwise}
    \end{cases}
\label{overall}
\end{equation}

\section{Experiments}
\label{experiments}
The agent is evaluated every 1,000 environment steps on 20 randomly selected instances. We use 100 * 1-box instances, 100 * 2-boxes instances and 60 * 3-boxes instances(since more boxes are more expensive, we use 60 instead of 100). The  results shown are averaged over 5 runs with different random seeds.

\begin{figure}[!ht]
    \centering
    \includegraphics[scale=0.8]{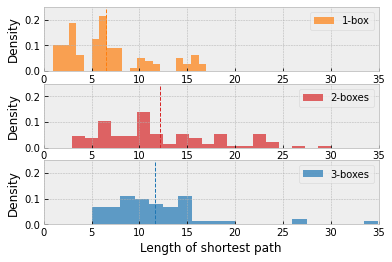}
    \caption{The shortest path of different Sokoban instances, dash lines are means. Top: solutions of 1-box instances, and the mean of solutions is 6.52. Mid: solutions of 2-boxes instances, and the mean of solutions is 12.14. Bottom: solutions of 3-boxes instances, and the mean of solutions is 11.62.}
    \label{fig:distribution}
\end{figure}

The length of the shortest path of an instance could indicate the difficulty of the instance. Fig.~\ref{fig:distribution} shows the distribution of shortest paths of instances we are using; as expected the more boxes, the longer the shortest path.

\begin{figure}[!ht]
    \centering
    \includegraphics[scale=0.3]{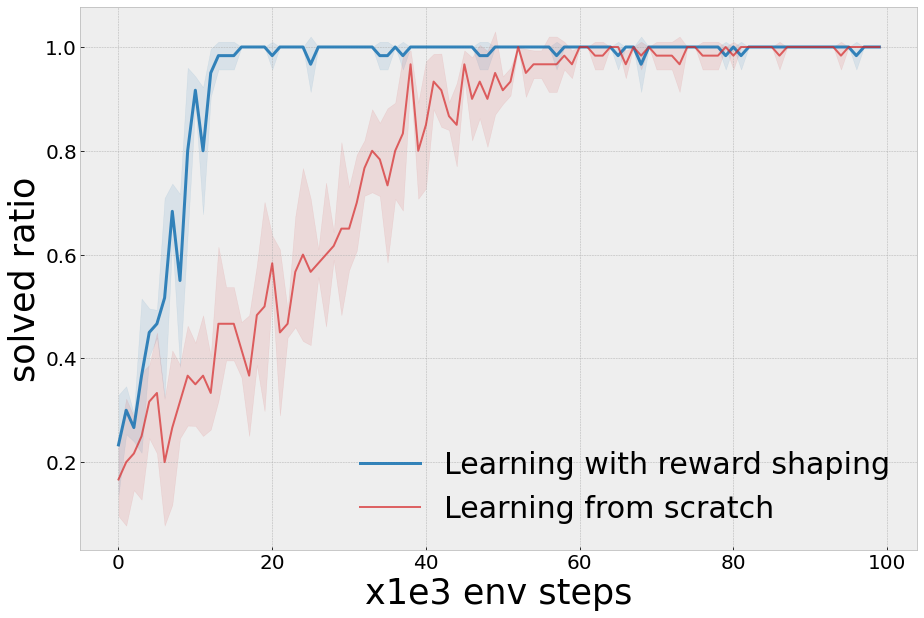}
    \caption{Learning over 100 * 1-box instances.}
    \label{fig:1}
\end{figure}

The learning results on 1-box instances is shown in Fig.~\ref{fig:1}. Even without reward shaping, the agent can quickly learn to master given instances within 60k environment steps. From the top subplot in Fig.~\ref{fig:distribution} we see that, solutions of 1-box instances are mostly shorter than 10. This also indicates that RL is able to solve simple sparse-reward problems. By adding reward shaping, the agent is about four times faster than learning from scratch. Both learning with reward shaping and learning from scratch are able to solve given instances within the given steps.

Learning on 2-box and 3-box instances is more difficult than learning on 1-box instances. Reinforcement learning from scratch almost learns nothing on 2-box and 3-box instances. Solutions of multiple box instances are generally longer than solutions of single-box instances.  Learning with reward shaping performs better by a large margin, the agent is able to solve given instances within 50k environment steps, while learning from scratch only reaches a solved ratio of around 0.2. This again demonstrates that RL is not good at solving sparse-reward problems.

\begin{figure}[!ht]
\begin{subfigure}{.5\textwidth}
  \centering
  \includegraphics[width=.99\linewidth]{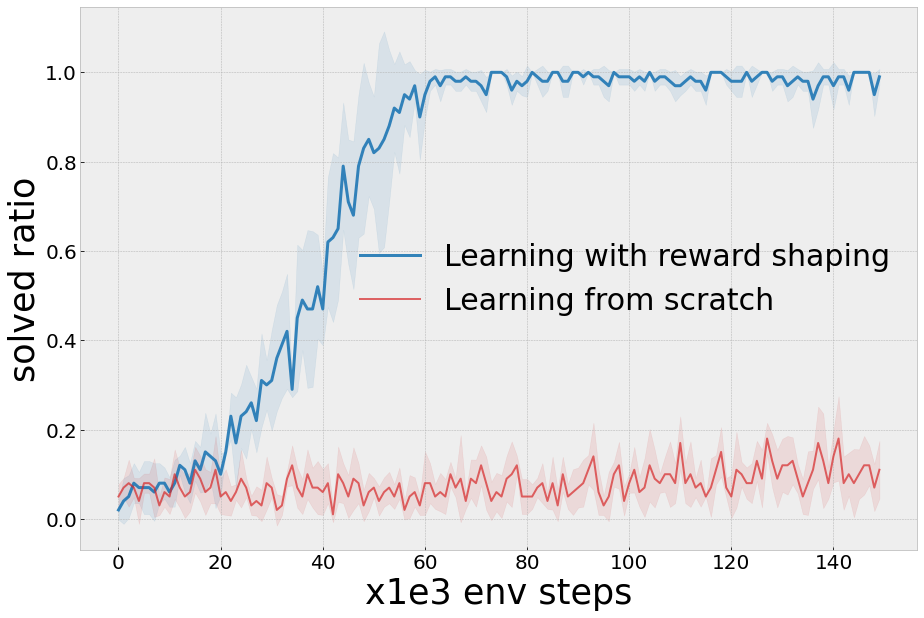}
\end{subfigure}
\begin{subfigure}{.5\textwidth}
  \centering
  \includegraphics[width=.99\linewidth]{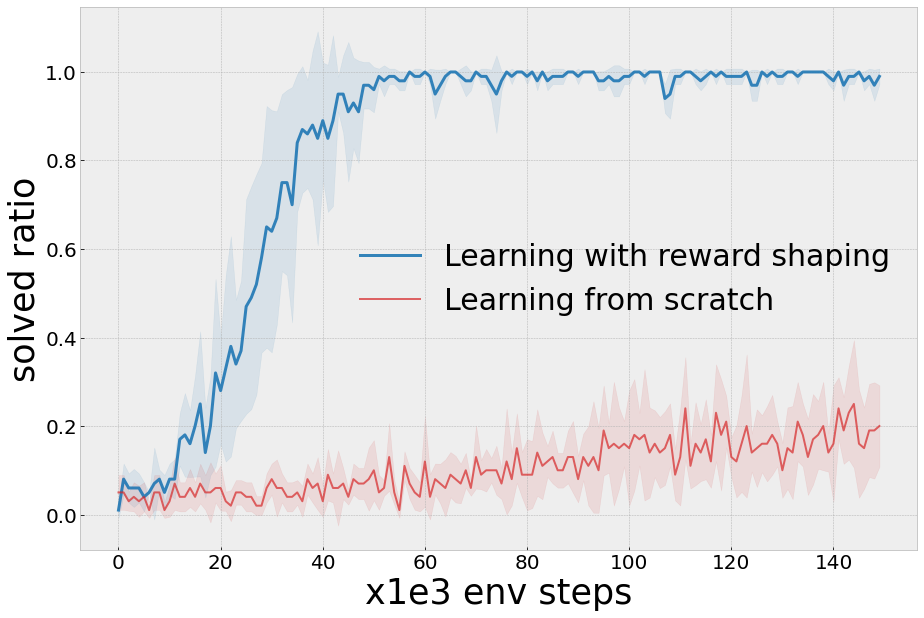}
\end{subfigure}
\caption{Learning over 100 * 2-boxes instances(left) and 60 * 3-boxes instances(right).}
\label{fig:2and3}
\end{figure}

\section{Discussion and Conclusion}
\label{conclusion}
In this work, we  showed that the distance function can be used as potential function in potential-based reward shaping to further speed up the learning in Sokoban. Meanwhile, we have seen that the distance function can be used as potential function in grid-world navigation tasks, since grid-world navigation tasks can be treated as special types of Sokoban where there are no needs for pushing boxes but only moving the agent to the target. Our experiments  showed that abilities of solving multiple instances can be quickly learned and compressed into a single behavior policy, which can be seen as the first step towards training a general policy which is able to solve unseen Sokoban instances quickly. For instance, if the agent is exposed to a Sokoban generator for training and the goal is to train the agent to be able to solve new unseen instances. Then learning with reward shaping will be way faster than learning from scratch to reach this 'final' goal.

A limitation of our approach is that since we are using search algorithms to provide the distance function, scalability is limited. For more difficult instances, search algorithms can not find solutions within a reasonable time, and the reward shaping we did in this work will not be usable. In the future, it would be interesting to work on scalable heuristic functions to use as potential functions in Sokoban. On the other hand, as we mentioned, our methods are the first used to train a baseline agent quickly, then reuse or transfer the trained neural networks for further training or tasks.

\section*{Acknowledgement}
The financial support to Zhao Yang is from the China Scholarship Council(CSC). Computation support is from ALICE/Leiden and the DSLab. The authors thank Michiel van der Meer, Hui Wang, Matthias M\"uller-Brockhausen and all members from the Leiden Reinforcement Learning Group for helpful discussions. Especially thanks to Thomas Moerland for detailed  feedback.

\bibliographystyle{splncs04}
\bibliography{main}

\newpage
\appendix
\section{Environment}
\label{env}
The environment we are using is from the implementation~\cite{SchraderSokoban2018} with slight modification. The rewards of the environment shown in the Tab.~\ref{tab:rewards}. Solving the instance by pushing all boxes onto targets returns 10.0; pushing one box onto a target gets 1.0 and pushing it off gets -1.0; in order to incentivize the agent solve the instance quickly, an -0.1 is given for each step the agent makes.
\begin{table}[!ht]
    \centering
    \begin{tabular}{|c|c|}
    \hline
        actions & reward \\
        \hline
         push all boxes on targets & 10.0\\
         \hline
         push one box onto target & 1.0\\
         \hline
         push one box onto target & -1.0\\
         \hline
         each step & -0.1 \\
         \hline
    \end{tabular}
    \caption{Rewards in the environment.}
    \label{tab:rewards}
\end{table}
\section{Neural Network Details}
\label{nn}
The model we are using contains three convolutional layers with kernel size 8x8, 4x4, 3x3, strides of 4, 2, 1, and number of output channels 32, 64, 64. Then followed by a fully connected layer with 512 units. In the end, the outputs are fed into two heads: outputting the policy logits and the state value. $\texttt{ReLU}$ is used as the activation function after each layer and \texttt{RMSprop} is the optimizer we used. The input is pixel image gets from the environment directly, which is 3x80x80. 
\begin{table}[!ht]
    \centering
    \begin{tabular}{|l|r|}
        \hline
        learning rate & $7\cdot 10^{-4}$\\
        \hline
        gamma & 0.99\\
        \hline
        entropy coef & 0.1\\
        \hline
        value loss coef & 0.5\\
        \hline
        eps & $10^{-5}$\\
        \hline
        alpha & 0.99\\
        \hline
        rollout storage size & 5\\
        \hline
        No. of environments for collecting trajectories & 30 \\
        \hline
    \end{tabular}
    \caption{Hyper-parameters of the neural network and training.}
    \label{tab:hp}
\end{table}

\section{Overall Training Loop}

\SetKwComment{Comment}{/* }{ */}
\begin{algorithm}[!ht]
\caption{Overall RL training loop}\label{alg:loop}
\SetKwInOut{Initialization}{Initialization}
\SetKwInOut{Output}{output}
\Initialization{policy $\pi$, number of training steps $N$, environment \texttt{env}}
$s$ $\gets$ \texttt{env.reset()}\;
\While{$n < N$}{
    $a$ $\gets$ $\pi(s)$\;
    $s'$, $r$, $\gets$ \texttt{env.step}($a$)\;
    \tcc{calculate the potential value under different situations.}
    \uIf{$s$ and $s'$ are solvable}
    {$f$ $\gets$ $-d(s') + d(s)$ \Comment*[r]{$f$ is the potential value.}}
    {\uElseIf{$s$ is solvable and $s'$ is not solvable}{
      $f$ $\gets$ $-d(s)-1$\;
    }
  }{\uElseIf{$s, s'$ are not solvable}{
    $f$ $\gets$ 0;
    }
  }
    $r'$ $\gets$ $r$ + $f$ \Comment*[r]{Reshape the reward.}
    execute A2C update on $\pi$ using the shaped reward $r'$\;
    $n \gets n+1$\;
}
return $\pi$ \;
\end{algorithm}

\end{document}